**Identifying Reasons for Contraceptive Switching from Real-World Data Using Large Language Models**

Brenda Y. Miao[1*], Christopher YK Williams[1], Ebenezer Chinedu-Eneh[2], Travis Zack[1,3], Emily Alsentzer[4,5], Atul J. Butte[1,6†], Irene Y. Chen[7,8,9†]

1. Bakar Computational Health Sciences Institute, University of California San Francisco, San Francisco, CA, USA
2. Department of Medicine, University of California San Francisco, San Francisco, CA, USA
3. Helen Diller Family Comprehensive Cancer Center, University of California San Francisco, San Francisco, CA, USA
4. Division of General Internal Medicine, Brigham and Women's Hospital; Boston, MA
5. Harvard Medical School, Boston, MA USA
6. Center for Data-driven Insights and Innovation, University of California, Office of the President, Oakland, CA
7. Computational Precision Health, University of California, Berkeley and University of California, San Francisco, Berkeley, CA, USA
8. Electrical Engineering and Computer Science, University of California, Berkeley, Berkeley, CA, USA
9. Berkeley AI Research, University of California, Berkeley, Berkeley, CA, USA

*Corresponding Author
Email: brenda.miao@ucsf.edu
Bakar Computational Health Sciences Institute, 490 Illinois Street
2nd Fl, North Tower, San Francisco, CA 94143

†Equal contribution

## Abstract

**Background:** Understanding why patients switch contraceptives is of significant interest but these factors are often only captured in unstructured clinical notes and can be difficult to extract. We evaluate the zero-shot abilities of a recently developed large language model, GPT-4, to identify reasons for switching between classes of contraceptives from clinical notes.

**Methods:** Clinical notes associated with contraceptive switches, defined as a difference in prescribed contraceptive modality between consecutive encounters, were selected from the UCSF Information Commons dataset. GPT-4 (HIPAA-compliant Microsoft Azure API) was used to extract contraceptive class started, stopped, and reasons for switching from these clinical notes. We evaluated six prompts on a subset of human annotated notes, and extracted reasons for contraceptive switching from the full dataset using the best prompt based on microF1 scores. BERTopic topic modeling was used to identify key reasons for contraceptive switching and enrichment of topics was assessed in patient subgroups stratified by race and ethnicity.

**Results:** We extracted 1,964 contraceptive switches from 1,515 patients. Younger patients and patients with self-reported race/ethnicities listed as "Black/African American" and "Latinx" had significantly higher rates of switching. GPT-4 demonstrated extraction of contraceptives started and stopped had MicroF1 scores of 0.849 and 0.881, respectively. Accuracy of GPT-4 extracted reasons for switching was 91.4% and 2.2% contained hallucinations. Key reasons for switching included patient preference, adverse events, and insurance. "Weight gain/mood change" and "insurance coverage" topics were enriched in patients self-reporting as being "Black or African American", "Latinx", or "Multi-Race/Ethnicity".

**Conclusion:** We demonstrate that GPT-4 can accurately extract reasons for contraceptive switching and identify certain reasons that are disproportionately found in specific demographic populations. Our approach to prompt development, and pipelines for extraction of reasons for medication switching, are also applicable to other classes of medication.

## Introduction

Prescription contraceptives play a critical role in supporting women's reproductive health. With the increasing availability of different contraceptives, providing patients with new options to manage their reproductive health, there is also a growing need to provide patients and providers with data-driven guidelines for informed decision-making[1–3]. Contraceptives may vary by active ingredient, with either progestin-only active ingredients or estrogen-progestin combinations available, as well as by mode of administration, which may be intrauterine (Intrauterine devices, IUDs), oral, transdermal, intravaginal, subdermal, or as an injection[4]. Each of these contraceptives have unique adverse event profiles that may contribute to clinical decision making[5]. In addition, several other factors, including personal preference, cost, availability, comorbidities and clinical constraints, may contribute to a patient's decision to start, stop, or switch contraceptives[6]. With nearly 50 million women in the United States using contraceptives[7], understanding the factors that drive contraceptives selection and switching is of significant interest.

Previous studies of medical claims data have shown that 44% of women starting a contraceptive discontinued its use within one year, although 76% resumed use of the same or another contraceptive within three months[8]. Some studies have begun to move beyond analysis of discontinuation rates and have included interviews[9] or social media data[10] to better understand the complexity of contraceptive switching. Using text mining approaches, these studies have identified specific patient subgroups that switch at different rates[11] or showed that patients have differing preferences and reasons for seeking contraceptives[4]. However, these studies may not capture the breadth or depth of clinical information found in medical record data and require development of custom machine learning models or time-consuming manual analysis to generate insight from the complexity of real-world text data[12–14].

Recently, the development of general large language models (LLMs) has shown significant promise in being able to extract medication information without the need for manually annotated training data ("zero-shot extraction")[12,15,16]. Despite concerns including factually incorrect information, clinicians and researchers remain optimistic that these computational advances can translate to clinically-meaningful use cases[17–20]. Here, we evaluate the ability of GPT-4 to extract contraceptive selection strategies and identify reasons for switching between classes of contraceptives using clinical notes from a large academic medical center.

## Materials and Methods

### Contraceptive switching cohort selection

A contraceptive switching cohort was selected from the UCSF Information Commons dataset[21], which contains deidentified structured data and clinical notes from over 6 million patients between 2012-2023. Clinical text notes were certified as deidentified as previously described[22] and are usable by UCSF researchers as non-human subjects research.

We identified all patients prescribed at least one contraceptive documented in the structured medication data based on a "therapeutic class" label. Non-drug contraceptives (e.g diaphragms/cervical caps, condoms, vaginal pH modulators, and spermicides), progestin and estrogen-containing agents not used for contraceptive purposes, and emergency contraceptives were removed (Table S1). The remaining contraceptives were mapped to the following modalities: Oral, Implant, Intrauterine device (IUD), Injection (intramuscular or subcutaneous), Transdermal, and Intravaginal based on regular expression values (Table S2). Contraceptives prescribed without a start date or associated clinical note and duplicate orders at each encounter date were removed. To filter out short notes without any relevant information, only clinical notes containing more than 50 tokens, created using encodings from OpenAI's open-source tokenizer tiktoken.

The dataset was further filtered to patients with encounters at least 6 months after the prescription of the first contraceptive, ensuring those without a switch weren't lost to follow-up. Prescriptions were sorted by documented start date, and encounters that contained a contraceptive switch were retrieved. A contraceptive switch was defined as a difference in prescribed contraceptive modalities between consecutive encounters.

Self-reported demographic information on race/ethnicity and preferred language were extracted from structured data, which was also used to calculate age at date of first contraceptive prescription. This study was conducted using retrospective, deidentified clinical data and was determined to be exempt from IRB review. All data were stored or processed on HIPAA compliant hardware at UCSF or through a HIPAA compliant Microsoft Azure instance ("UCSF Versa"). No data was transferred or stored by OpenAI; and OpenAI settings were maintained so that no prompt information would be stored, even temporarily.

**Prompt evaluation for extraction of contraceptive selection strategy**
Prompting can have significant effects on the accuracy of large language models[23,24]. We tested six prompts (Table S3), varying both system information and output formats, to extract the following information: 1) which contraceptive was stopped, 2) which new contraceptive was started, and 3) why the contraceptive switch occurred. To avoid overfitting, these six prompts were evaluated on a held-out subset of contraceptive switching clinical notes from 5% of the patients. The model used was GPT-4, with temperature set at 0, maximum response length capped at 500 tokens, top_p set to 1, and all other parameters kept as default. A zero-shot approach was used, with no additional information or training data provided outside of the encounter's associated clinical note. Resulting values were mapped to the six contraceptive modalities using regular expression values (Table S2). All GPT-4 queries were performed between November 13-15, 2023.

A clinical evaluator (EE) assessed the accuracy of GPT-4 extraction for contraceptives started and stopped within each note. Micro F1 scores, which represent the harmonic mean of precision and recall scores, are reported. The best prompt was selected based on the highest average score attained across all medications started/stopped determined by manual evaluation. The clinical reviewer was also instructed to identify whether the extracted reason was accurate based on the clinical note and whether any hallucination occurred, which was defined as information produced by the language model that could not be derived from the clinical note.

**Comparison of GPT-4 contraceptive information extraction to baseline models**
The best prompt selected from the development dataset was applied to the remaining 95% "test set" of the contraceptive switching cohort using the same GPT-4 setup. We compared our LLM-based methods against several traditional machine learning techniques, including logistic regression, random forest, and BERT-style models. Since human clinical annotations were not available for this larger dataset, weak labels from structured data, specifically which contraceptives were started and stopped at the associated clinical encounter, were used for training and evaluation in each of these models. Structured data may not reflect the contents of clinical notes if patients are prescribed contraceptives at a different facility or stop dates are not documented, so we compared these silver-standard labels to human annotation for the 93 clinical notes in the prompt evaluation set using Cohen's Kappa coefficient to assess reliability between the two sources.

Two sets of logistic regression and random forest models were developed using either bag-of-words and term-frequency inverse document frequency (TF-IDF)[25] text representations. Multiclass classification was performed, with models predicting the modality of contraceptives started or stopped (oral, IUD, subdermal, intravaginal, injection, transdermal). We performed 5-fold cross validation using a 70/10/20 split between train, validation, and test data. Due to differences in training sizes between baseline models and GPT-4, this split is independent of the previous prompt evaluation and GPT-4 test sets. Hyperparameter tuning was performed using a grid search of varying regularization values (C=[0.01, 0.1, 1, 10, 100, 1000]) for logistic regression and both number of estimators and max depth for random forest (n_estimators=[50, 100, 250, 500], max_depth=[20, 50, 100]).

The UCSF-BERT model[26,27] trained on a large corpus of clinical notes was also used as a baseline. Again, we performed 5-fold cross validation using a 70/10/20 split. Hyperparameter tuning was performed using Optuna[28], and both learning rate and weight decay were varied (learning rate=(1e-5, 5e-5), weight decay=(4e-5, 0.01)). Models were trained for 5 epochs, with early stopping. To accommodate for the 512 maximum token length allowed by UCSF BERT, a sliding window was used with final prediction selected by majority vote across all windows.

To simulate few-shot learning, we trained each of the baseline models on random subsamples of 100%, 50%, 25%, 10%, 5%, and 1% of the training data. Micro-averaged F1 scores are reported for each model on the held-out test set.

**Unsupervised clustering of extracted reasons for contraceptive switching**
GPT-4 was also used to extract reasons for contraceptive switching from the test set using the best prompt. To identify key reasons for medication switching, we applied BERTopic, a topic modeling method that clusters document embeddings, to all reasons extracted from both the prompt evaluation and test sets. The UCSF-BERT model was used to generate embeddings from the list of extracted reasons and embeddings were clustered by BERTopic[29]. Briefly, dimensionality reduction was applied to the embeddings using Uniform Manifold Approximation and Projection (UMAP), with 5 components and 3 neighbors with euclidean distance metrics. HDBSCAN[30] was used to cluster reduced embeddings, with number of topics dynamically chosen by the algorithm, and TF-IDF used to identify key terms from each cluster. All other default parameters were used. Topics were manually reviewed and similar topics were grouped together.

Subgroup analysis was performed to understand whether topics were associated with particular patient demographics. Adapting from previous enrichment methods[31], we used topic probabilities assigned to each document by the BERTopic model to calculate a weighted enrichment score that describes the relative contribution of each topic to patient subgroups. Specifically, enrichment scores were calculated as $\theta_{k,j} = \frac{q_{n,k} \cdot y_{n,j}}{\sum_{n=1}^{N} q_{n,k} * \sum_{n=1}^{N} y_{n,j}}$, where q(n,k) describes the weight of each topic k for note n, and y(n,j) are the patient subgroups assigned to each note. The scores were normalized by total topic weight, as well as by number of patients in each subgroup, and reported scores were negative log transformed.

**Statistics**
We present means and standard deviations for continuous distribution, and utilize student t-tests to analyze differences in continuous distributions. To evaluate differences in categorical data, Chi-square tests were applied. Statistical analyses were conducted using the SciPy package[32], and a p-value less than 0.05 was used to indicate statistical significance.

## Results

**Contraceptive switching cohort**

We selected a contraceptive patient cohort using the UCSF Information Commons dataset. We identified 37,834 patients with at least 1 medication order for an intrauterine, oral, intravaginal, subdermal, transdermal, or injectable contraceptive. We removed 5,594 patients who did not have any follow up encounters at least 6 months after the last contraceptive order. This left 37,834 patients with 100,593 medication orders. We further filtered out the 11,916 orders without associated clinical notes and 53,125 duplicate medication orders, leaving a contraceptive cohort consisting of 39,790 medication orders across 20,283 unique patients (Figure 1).

Among this contraceptive cohort, 1,515 (7.6%) patients experienced a total of 1,964 contraceptive switches. Compared to patients who did not have a contraceptive switch, patients with contraceptive switches tended to be younger, with a mean age of 25.9 years (SD: 7.7) compared to 29.1 years (SD: 8.4, p<0.001, Table 1). There was also a statistically significant difference in the proportion of patients with and without contraceptive switches by patient race/ethnicity (p<0.001). The largest difference occurred in patients with a race/ethnicity listed as "Black or African American," with 19.3% of such patients having a contraceptive switch compared to 8.2% without. Patients identifying as "Latinx" were also more likely to have a contraceptive switch (19.3%) compared to the proportion of "Latinx" patients without contraceptive switches (15.1%). "White" (33.0%) or "Asian" (16.0%) patients had lower rates of contraceptive switching in this cohort compared to the same groups without switches, with 45.1% of patients without contraceptive switches identifying as "White" and 20.3% identifying as "Asian."

Switching differed significantly by the first contraceptive prescribed, with the highest rates of switching following initial prescription of transdermal contraceptives (33.5%) and the lowest rates following initial prescription of intrauterine (5.1%) and oral (6.3%) contraceptives. The most common switch occurred in patients who were on oral contraceptives and switched to intravaginal contraceptives (n=205, n=10.5%). The least common switch occurred from intrauterine to injectable contraceptives (n=6, 0.31%, Table S4).

**Human evaluation of GPT4 extraction of contraceptive switching**

Prompt evaluation was performed on a held out set consisting of notes from 5% of patients (n=93 clinical notes), and evaluated against annotations from a clinical reviewer. There was no significant difference in performance across the six prompts used to extract contraceptive information using zero-shot GPT-4, with micro F1 scores ranging from 0.817 to 0.849 (mean=0.827, SD: 0.012) for extraction of contraceptive started, and 0.827 to 0.881 (mean=0.854, SD: 0.020) for extraction of contraceptive stopped (Figure 2). The best prompt for medication stopping extraction used the specialist system configuration and default prompt. Reasons extracted by this prompt were also evaluated by a clinical reviewer for both accuracy and rate of hallucination. Human evaluation showed that GPT-4 was capable of extracting these reasons with 91.4% accuracy and without hallucination 97.8% of the time (n=93). Given the high accuracy and minimal hallucination of this prompt for extracting information about contraceptive stopping and reasons for stopping on the development dataset, this prompt was selected to extract contraceptive information from the remaining clinical notes.

**GPT-4 contraceptive switching information extraction outperforms baseline models**
Zero-shot GPT-4 performance using the best prompt was also compared to baseline models trained on different proportions silver-standard labels derived from structured data. GPT-4 outperformed all baseline models, regardless of the proportion of training data used for baseline models (Figure 3, with micro F1 scores of 0.828 and 0.439 on contraceptive start and stop extraction, respectively. The next best model was random forest trained on TF-IDF representations, with a 0.714 (SD: 0.024) score on medication start and 0.424 (SD: 0.009) on medication stopping.

Concordance between silver-standard labels and human annotations available showed a Cohen's Kappa coefficient of 0.585 for medication starting labels and 0.217 for contraceptive stopping (n=93). When we removed notes without relevant contraceptives, determined by the human evaluator, concordance between these two methods increased to 0.960 for contraceptives started and 0.644 for contraceptives stopped (Table S5).

**Identification of reasons for contraceptive switching**
Unsupervised BERTopic topic modeling of extracted reasons for stopping across the full dataset identified 19 topics, which were manually grouped into 10 cohesive topics (Table S6). Excluding the 1136 notes that did not contain a relevant reason (topic 0, Table S7), the most frequently occurring topics contained terms related to spotting and irregular bleeding (topic 1), desire to switch contraceptives (topic 2), and forgetting to take daily pills (topic 3). Topics 4, 6, 7 described other adverse events of contraceptive use, including irritation and rash, weight gain and mood changes, and irregular menses and pain. Topic 5 related to IUD malpositioning and removal, and topic 9 related to implant removal. Finally, topic 8 included terms related to insurance coverage ([Figure 4A](Figure 4A)).

Subset analysis stratified by race/ethnicity identified enrichment of specific topics within certain patient subgroups. Weight gain and mood change (topic 6) were enriched in patients who self-reported as being "Latinx" or "Other", and showed lower enrichment in patients self-reporting as "Black or African American". Topic 9 (Implant removal) was enriched in patients who self-reported a race/ethnicity of "Asian", and topic 8 (insurance coverage) was enriched in patients of "Black or African American", "Latinx", or "Multi-Race/Ethnicity" (Figure 4B).

## Discussion

We demonstrated that GPT4 can accurately extract which medications were started and stopped during an encounter from associated clinical notes. GPT-4 performance, evaluated by both automated analysis and gold-standard manual annotation, was stable between six different prompts although more complex prompting methods may further improve medication information extraction[24,33]. We further showed that the majority of reasons for contraceptive switching extracted by GPT-4 were also correct, with minimal hallucinations.

Lastly, we uncovered latent contraceptive-specific reasons for switching medications by clustering embeddings derived from GPT-4 extracted values. Topic clusters ranged from treatment failure to patient preference, as well as adverse events and insurance reasons. In line with previous studies[34], we showed that weight gain and mood changes as reasons for switching were enriched in patient populations who self-reported their race/ethnicity as "Latinx" or "Other". Additionally, we showed that insurance coverage as a reason for switching disproportionately affected patients identifying as "Latinx" or "Black or African American." Our results highlight recent concerns regarding financial barriers to contraceptive access and resulting racial inequities in reproductive health[35]. Future validation in independent datasets will be needed to determine whether this difference persists in larger samples or in other cases of medication switching.

There are several limitations to this study. This dataset is limited to values derived from a large, academic medical center, which may introduce bias in the types of patients or contraceptives captured. We assume that clinical notes contain information on all medications ordered at the same or previous encounters, but some medications may not be discussed or documented. This is reflected in poor concordance between structured data labels and human evaluation, particularly for medication stopping values. Additionally, because the de-identification process is not perfect, manual review of some notes identified several medication names that were inappropriately redacted. This was particularly prevalent among contraceptive brand names that resemble common patient names (eg. "Camila" or "Heather") that are deliberately redacted. Finally, another limitation of our work surrounds interpretability of results, which is of significant importance to clinical care. There is little public information provided about GPT4's training data, approach, or model architecture, and we have not yet tested any open-source language models on this task. As a result, we refrain from making conclusions about why LLMs like GPT-4 produces certain results, and focus instead on evaluating overall performance and insights that can be derived from extraction of information from clinical notes.

In conclusion, our findings demonstrate that reasons for contraceptive switching are disproportionately found in specific patient demographics. Our approach can be applied towards treatment strategy analysis across or within different classes of medications beyond contraceptive switching to improve understanding of treatment strategy and shape more detailed treatment effect estimation models.

## Data availability

To protect patient privacy, the clinical notes from this study are not made publicly available. All code to reproduce the methods and supplemental data described here are made available at https://github.com/BMiao10/contraceptive-switching

## Author contributions

Conceptualization: BYM, AJB, IYC; Data curation: BYM, EC; Software/Formal analysis: BYM, CYKW, EC; Methodology: BYM, CYKW, TZ, IYC; Supervision: AJB, IYC; Writing – original draft preparation: BYM, EA, IYC; Writing – review & editing: BYM, CYKW, TZ, EA, AJB, IYC.

## Disclosures

**BYM** has no conflicts of interest to disclose. **CYKW** has no conflicts of interest to disclose. **EC** has no conflicts of interest to disclose. **TZ** has no conflicts of interest to disclose. **EA** reports personal fees from Canopy Innovations, Fourier Health, and Xyla; and grants from Microsoft Research. **IYC** is a minority shareholder in Apple, Amazon, Alphabet, and Microsoft. **AJB** is a co-founder and consultant to Personalis and NuMedii; consultant to Samsung, Mango Tree Corporation, and in the recent past, 10x Genomics, Helix, Pathway Genomics, and Verinata (Illumina); has served on paid advisory panels or boards for Geisinger Health, Regenstrief Institute, Gerson Lehman Group, AlphaSights, Covance, Novartis, Genentech, and Merck, and Roche; is a shareholder in Personalis and NuMedii; is a minor shareholder in Apple, Facebook, Alphabet (Google), Microsoft, Amazon, Snap, 10x Genomics, Illumina, CVS, Nuna Health, Assay Depot, Vet24seven, Regeneron, Sanofi, Royalty Pharma, AstraZeneca, Moderna, Biogen, Paraxel, and Sutro, and several other non-health related companies and mutual funds; and has received honoraria and travel reimbursement for invited talks from Johnson and Johnson, Roche, Genentech, Pfizer, Merck, Lilly, Takeda, Varian, Mars, Siemens, Optum, Abbott, Celgene, AstraZeneca, AbbVie, Westat, and many academic institutions, medical or disease specific foundations and associations, and health systems. Atul Butte receives royalty payments through Stanford University, for several patents and other disclosures licensed to NuMedii and Personalis. Atul Butte's research has been funded by NIH, Peraton (as the prime on an NIH contract), Genentech, Johnson and Johnson, FDA, Robert Wood Johnson Foundation, Leon Lowenstein Foundation, Intervalien Foundation, Priscilla Chan and Mark Zuckerberg, the Barbara and Gerson Bakar Foundation, and in the recent past, the March of Dimes, Juvenile Diabetes Research Foundation, California Governor's Office of Planning and Research, California Institute for Regenerative Medicine, L'Oreal, and Progenity. None of these organizations or companies had any influence or involvement in the development of this manuscript.

## Funding

Research reported in this publication was supported by the National Center for Advancing Translational Sciences, National Institutes of Health, through UCSF-CTSI Grant Number UL1 TR001872. The content is solely the responsibility of the authors and does not necessarily represent the official views of the National Institutes of Health.

**Acknowledgements**

The authors acknowledge the use of the UCSF Information Commons computational research platform, developed and supported by UCSF Bakar Computational Health Sciences Institute in collaboration with IT Academic Research Services, Center for Intelligent Imaging Computational Core, and CTSI Research Technology Program. We also thank Madhumita Sushil for valuable feedback on this project.

# Figures

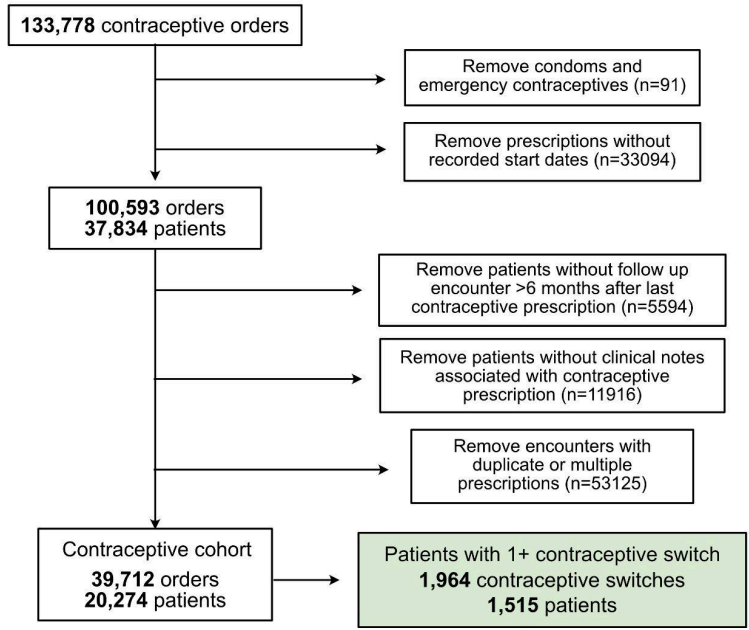 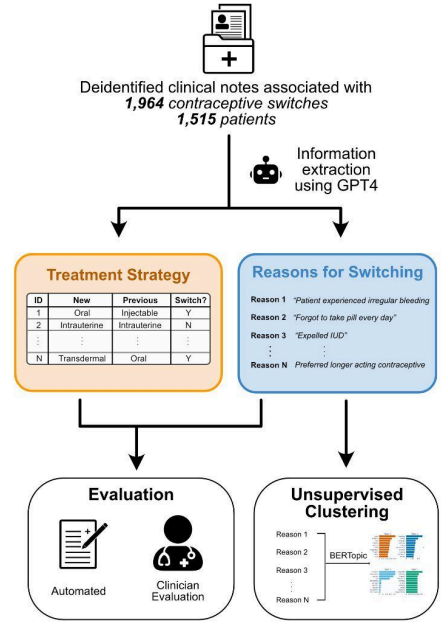

**Figure 1**: Study overview. A) We selected a contraceptive patient cohort from the UCSF Information Commons dataset. Among 20,283 patients with unique contraceptive prescriptions and associated clinical notes, 1,515 (7.6%) patients experienced a total of 1,964 total contraceptive switches. B) Study overview to assess the ability for GPT4 to extract contraceptive switching values from clinical notes, and to identify key reasons for switching using unsupervised clustering methods.

**Table 1**: Contraceptive prescription cohort demographics

|  | **Contraceptive switch (n=1515)** | **No switch (n=15,907)** | **Significance** |
|---|---|---|---|
| **Age, mean (SD)** | 25.9 (7.7) | 25.9 (7.7) | p<0.001 |
| **Race/Ethnicity, n (%)** | | | p<0.001 |
| *Missing (n)* | *32* | *815* | |
| White | 490 (33.0%) | 6813 (45.1%) | |
| Black or African American | 286 (19.3%) | 1237 (8.2%) | |
| Latinx | 286 (19.3%) | 2281 (15.1%) | |
| Asian | 237 (16.0%) | 3071 (20.3%) | |
| Other | 115 (7.8%) | 1224 (8.1%) | |
| Multi-Race/Ethnicity | 69 (4.7%) | 466 (3.1%) | |
| **Preferred Language, n (%)** | | | p<0.001 |
| *Missing (n)* | *0* | *5* | |
| English | 1474 (97.3%) | 15405 (96.9%) | |
| Spanish | 14 (0.9%) | 281 (1.8%) | |
| Other | 27 (1.8%) | 216 (1.4%) | |
| **First contraceptive prescribed, n (%)** | | | p<0.001 |
| Oral | 661 (43.6%) | 10496 (66.0%) | |
| Intravaginal | 244 (16.1%) | 1935 (12.2%) | |
| Injectable | 199 (13.1%) | 853 (5.4%) | |
| Transdermal | 187 (12.3%) | 558 (3.5%) | |
| Implant | 160 (10.6%) | 799 (5.0%) | |
| Intrauterine | 64 (4.2%) | 1266 (8.0%) | |

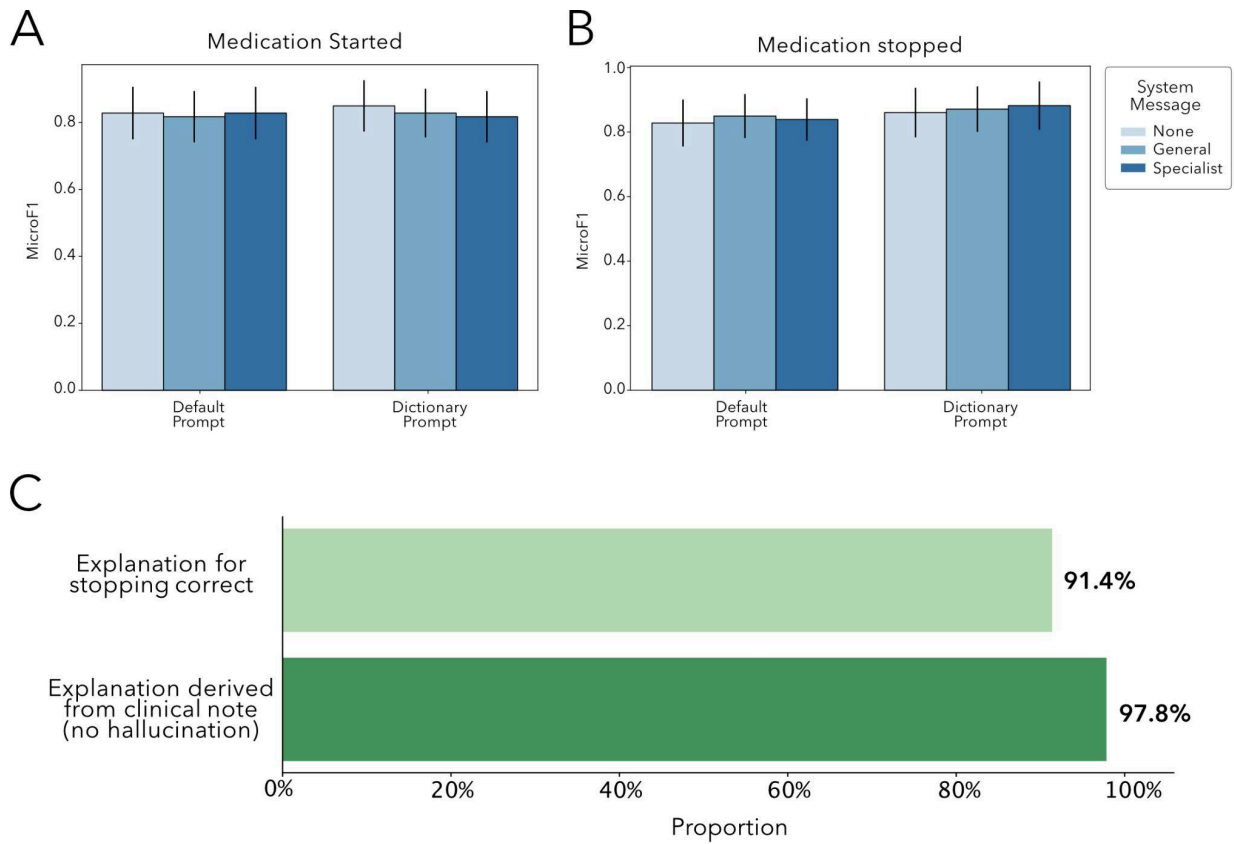

**Figure 2.** Development of prompt to extract contraceptive switching information. GPT4-extracted values for contraceptive class A) started and B) stopped compared to human annotation (n=93). C) Human evaluation was also performed to assess whether GPT-4 extracted reasons for contraceptive switching was accurate, and contained only information specifically mentioned in the associated clinical note (not hallucination).

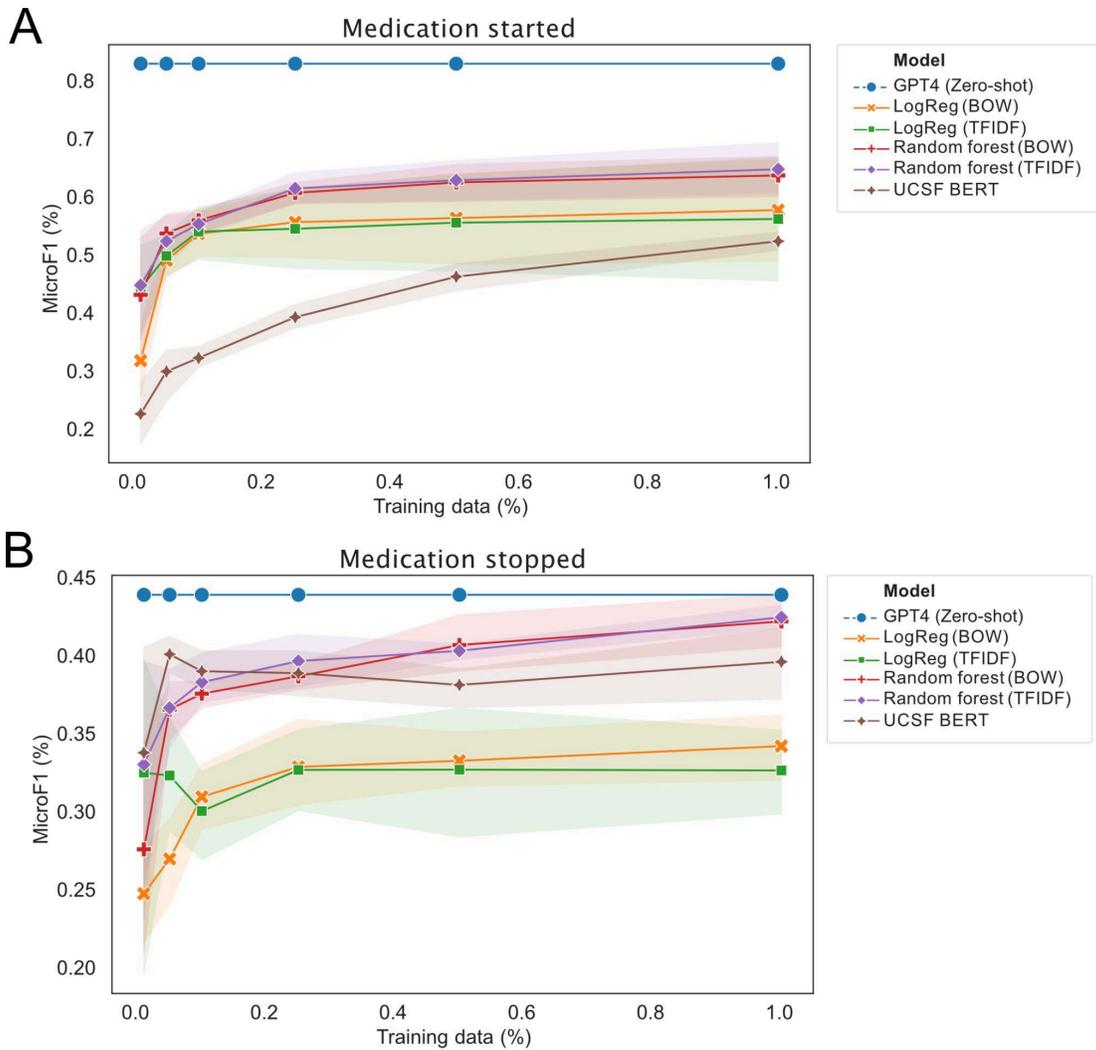

**Figure 3**: GPT-4 performance compared to baseline. Following prompt evaluation, GPT-4 performance on the remaining test set was also compared to baseline model performance for extraction of contraceptive A) started and B) stopped. Silver-standard labels from structured data were used for training and evaluation of baseline models, and for evaluation of zero-shot GPT-4.

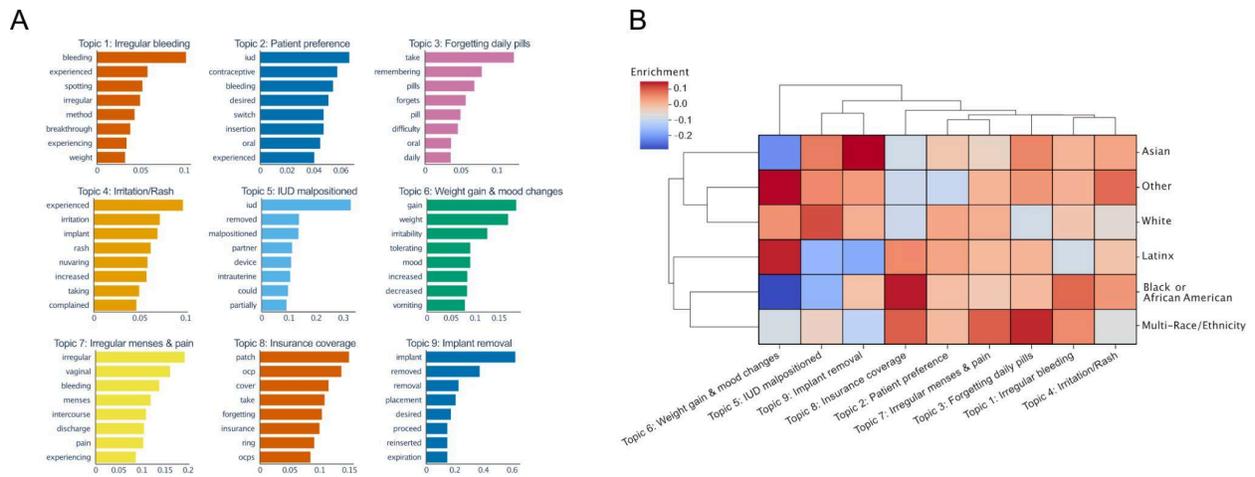

**Figure 4**: Clustering reasons for contraceptive switching using BERTopic. A) BERTopic modeling was used to cluster GPT-4 extracted reasons for contraceptive switching, with nine key topics identified. Top terms for each cluster are shown. B) Topics were assessed for enrichment amongst patient subgroups by race/ethnicity. Higher enrichment scores indicate higher prevalence of a topic written in notes within a patient subgroup.